\documentclass[journal]{IEEEtran}

\date{} 

\usepackage{enumitem}

\usepackage{url}

\usepackage{hyperref} 

\usepackage{amsmath,graphicx,caption,epstopdf,float}
\usepackage[usenames, dvipsnames]{color}

\ifCLASSINFOpdf
\else
\fi

\begin{document}

\onecolumn 

\begin{description}[labelindent=1.5cm,leftmargin=4.5cm,style=multiline]

\item[\textbf{Citation}]{T. Alshawi, Z. Long, and G. AlRegib, ``Unsupervised uncertainty estimation using spatiotemporal cues in video saliency detection,'' IEEE Transactions on Image Processing, vol. 27, no. 6, pp. 2818-2827, Jun. 2018.}
\\
\item[\textbf{DOI}]{\url{https://doi.org/10.1109/TIP.2018.2813159}}
\\
\item[\textbf{Review}]{Date of publication: 7 March 2018}
\\
\item[\textbf{Data and Codes}]{\url{https://ghassanalregibdotcom.files.wordpress.com/2018/02/tariq_tip2018_code.zip}}
\\
\item[\textbf{Bib}] {@article\{alshawi2018unsupervised,\\
  title=\{Unsupervised Uncertainty Estimation Using Spatiotemporal Cues in Video Saliency Detection\},\\
  author=\{Alshawi, Tariq and Long, Zhiling and AlRegib, Ghassan\},\\
  journal=\{IEEE Transactions on Image Processing\},\\
  year=\{2018\},\\
  volume=\{27\},\\
  number=\{6\},\\
  pages=\{2818-2827\},\\
  month=\{June\},\\
  publisher=\{IEEE\} \}
} 
\\

\item[\textbf{Copyright}]{\textcopyright 2018 IEEE. Personal use of this material is permitted. Permission from IEEE must be obtained for all other uses, in any current or future media, including reprinting/republishing this material for advertising or promotional purposes, creating new collective works, for resale or redistribution to servers or lists, or reuse of any copyrighted component of this work in other works.}
\\
\item[\textbf{Contact}]{\href{mailto:zhiling.long@gatech.edu}{zhiling.long@gatech.edu}  OR \href{mailto:alregib@gatech.edu}{alregib@gatech.edu}\\ \url{https://ghassanalregib.com/} \\ }
\end{description}

\thispagestyle{empty}
\newpage
\clearpage
\setcounter{page}{1}

\twocolumn

\title{Unsupervised Uncertainty Estimation Using Spatiotemporal Cues in Video Saliency Detection}

\author{Tariq Alshawi, Zhiling Long, and Ghassan AlRegib
\thanks{Copyright (c) 2018 IEEE. Personal use of this material is permitted.
However, permission to use this material for any other purposes must be
obtained from the IEEE by sending a request to pubs-permissions@ieee.org
\par Tariq Alshawi is with the Electrical Engineering Department, College of Engineering, King Saud Univeristy, Riyadh, Saudi Arabia, and also with the Center for Signal
and Information Processing (CSIP), School of Electrical and Computer Engineering,
Georgia Institute of Technology, Atlanta, GA, 30332 USA (e-mail: talshawi@
gatech.edu).
\par Zhiling Long and Ghassan AlRegib are with the Center for Signal
and Information Processing (CSIP), School of Electrical and Computer Engineering,
Georgia Institute of Technology, Atlanta, GA, 30332 USA (e-mail: zhiling.long@gatech.edu; alregib@gatech.edu).}}


\maketitle
\begin{abstract}
In this paper, we address the problem of quantifying the reliability of computational saliency for videos, which can be used to improve saliency-based video processing algorithms and enable more reliable performance and objective risk assessment of saliency-based video processing applications. Our approach to quantify such reliability is two fold. First, we explore spatial correlations in both the saliency map and the eye-fixation map. Then, we learn the spatiotemporal correlations that define a reliable saliency map. We first study spatiotemporal eye-fixation data from the public CRCNS dataset and investigate a common feature in human visual attention, which dictates a correlation in saliency between a pixel and its direct neighbors. Based on the study, we then develop an algorithm that estimates a pixel-wise uncertainty map that reflects our supposed confidence in the associated computational saliency map by relating a pixel's saliency to the saliency of its direct neighbors. To estimate such uncertainties, we measure the divergence of a pixel, in a saliency map, from its local neighborhood. Additionally, we propose a systematic procedure to evaluate uncertainty estimation performance by explicitly computing uncertainty ground truth as a function of a given saliency map and eye fixations of human subjects. In our experiments, we explore multiple definitions of locality and neighborhoods in spatiotemporal video signals. In addition, we examine the relationship between the parameters of our proposed algorithm and the content of the videos.  The proposed algorithm is unsupervised, making it more suitable for generalization to most natural videos. Also, it is computationally efficient and flexible for customization to specific video content. Experiments using three publicly available video datasets show that the proposed algorithm outperforms state-of-the-art uncertainty estimation methods with improvement in accuracy up to 63\% and offers efficiency and flexibility that make it more useful in practical situations.
\end{abstract}
\IEEEpeerreviewmaketitle
\section{Introduction}
\par Human visual attention modeling and understanding can be a key contributor to the advancement in computational analysis of big visual data which might offer similar computational efficiency to that of the human vision system (HVS). Algorithms for object detection and recognition \cite{Ren2014saliencyRecognition}, scene understanding \cite{Bharath2013}, video compression \cite{Gitman2014SAVAM}, and multimedia summarization \cite{peng2009keyframe} can be designed to exploit human visual attention mechanisms, and potentially, produce faster and more perceptually satisfying results. Driven by the fast responsiveness of HVS to low-level visual features, bottom-up spatiotemporal saliency detection has been crafted to identify perceptually unique objects in videos, and in turn predict the likelihood that a human would focus on these objects as opposed to the rest of the scene \cite{Itti05}.

\par
\begin{figure}
\centering
\captionsetup{justification=centering}
\includegraphics[width=0.5\textwidth]{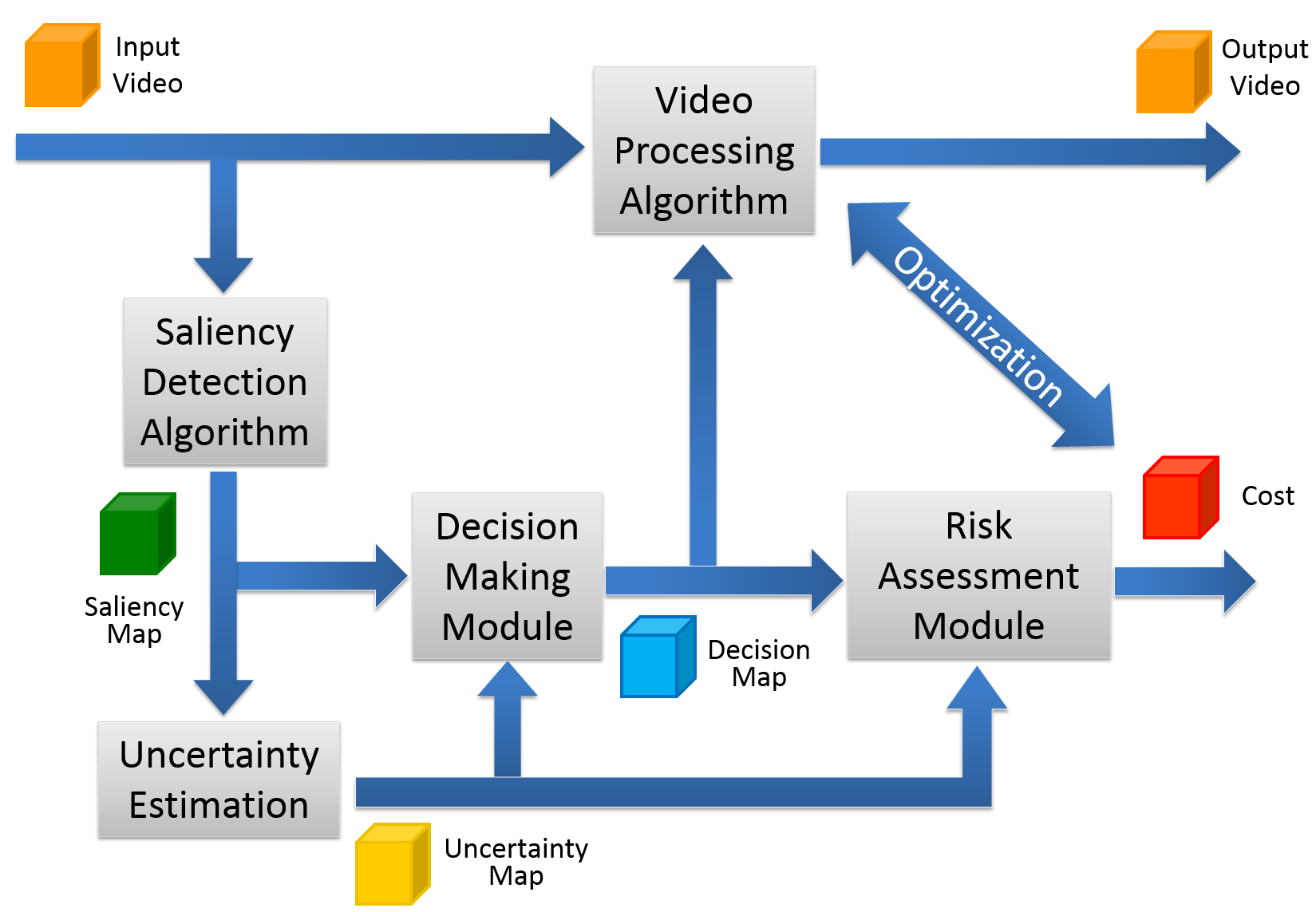}
\caption{Uncertainty-based framework for improving saliency-enabled video processing algorithms}
\label{fig:UncertFramework}
\end{figure}

 The majority of existing research efforts focus on computational saliency models  \cite{WangTIP2014} \cite{KimTIP2015} \cite{HuangTIP2014} \cite{MahapatraJSTSP2014} \cite{LiuTCSVT2014} \cite{LiangTIP2015}, however, less attention has been given to quantifying the reliability of the generated saliency maps \cite{Fang2014video} \cite{alshawi15temporal} \cite{alshawi15spatial}. The validity of such maps is crucial for integrating visual attention in various image and video processing applications.  It is a common practice to consider the validity of a saliency detection model, at every pixel, to be directly related to its average performance on image and video datasets. In other words, a saliency detection model is, first, evaluated using typical visual stimuli datasets with eye tracking data such as CRCNS \cite{CRCNS05}, MSRA \cite{Liu2011MSRA}, MIT \cite{Judd2009MIT}, and SAVAM \cite{Gitman2014SAVAM}. Then, algorithms that detect salient regions effectively, according to a predefined ground truth in the dataset, are assumed to perform well when used in various applications. However, such saliency detectors might fail to produce reliable results in certain contexts or situations, despite their superior performance in other contexts. Thus, it is important to consider the reliability of a saliency map given the context of the image or video at hand. Additionally, explicitly quantifying the reliability of a saliency map enables effective decision making and risk assessment in applications that exploit visual attention mechanisms. We recently proposed an uncertainty-based saliency-enabled framework \cite{alshawi15spatial}, shown in Fig.\ref{fig:UncertFramework}, for image and video processing applications that incorporates the reliability (uncertainty) of the computed saliency maps  to enable a systematic decision making process and facilitate risk assessment that depends on the application at hand such as object detection and recognition \cite{Ren2014saliencyRecognition}, scene understanding \cite{Bharath2013}, video compression \cite{Gitman2014SAVAM}, and multimedia summarization \cite{peng2009keyframe}.

\par Recently, there has been some research on uncertainty specific to image and video processing applications. In \cite{GuIETvision2015}, authors proposed using an active learning algorithm based on one-versus-one (OVO) strategy support vector machine (SVM) to solve multi-class image classification. The results of OVO SVM are combined according to a cost function that maximizes the diversity of the chosen set of examples and minimizes the uncertainty of the classification of this set. The uncertainty in this work is estimated using the difference in number of votes between the highest votes class and the second highest class. As the difference in number of votes increases, it is more likely that the highest votes class is the true representative class, so the uncertainty is lower. In the context of medical image registration, Saygili et al. \cite{SaygiliTMI2016} proposed a confidence measure that reflects the accuracy of the registration process of a pair of images. The proposed measure relates the confidence of the registration process at each pixel to the global minima and the steepness of a predefined cost function. The registration at a given pixel is expected to be more reliable if the associated cost function produces a global minima at that location and the cost function in its local region is very steep. In the context of stereo vision and depth estimation, numerous confidence measures have been proposed in literature \cite{HuTPAMI2012}. Typically, these measures associate the confidence of pixel's match with the shape of the matching cost function, e.g. sum of absolute differences, around that pixel. Haeusler et al. \cite{HaeuslerCVPR2013} proposed applying random decision forest framework on a large set of diverse stereo confidence measures to improve the performance of stereo solvers. 

\par In the context of saliency detection, there has been very limited work to address the problem of quantifying uncertainty. Directly applying uncertainty and confidence measures proposed for other image and video processing applications might not take into consideration characteristics of human visual attention mechanisms which are crucial for saliency detection. It is important to note here that the term uncertainty has been used in saliency detection research and visual attention modeling to describe a phenomenon that steers attention. The authors in \cite{feldman2010attention} argue that attention can be understood as inferring the level of uncertainty during perception. Other papers such as \cite{bruce2006saliency, bruce2009saliency, wei2014biologically, wang2010measuring} have proposed saliency attention models that are based on entropy and information theory measures that quantify the level of uncertainty in visual stimuli. However, in the context of our work, uncertainty estimation is mainly concerned with quantifying the reliability of saliency maps. In other words, we are interested in quantifying the confidence in saliency maps during decision making process rather than the entropy caused by uncertainty during perception. Relevant to this kind of uncertainty, the authors in \cite{Fang2014video} proposed a supervised method to estimate the uncertainty associated with detected saliency of a video pixel. The method uses binary entropy function to measure uncertainty according to the probability of a pixel being salient given the distance of the target pixel from the center of mass $p(s|d)$, and connectedness of the target pixel $p(s|c)$. The coordinates of the center of mass of saliency map $[x_c,y_c]$ are first calculated using the ground truth map. Then, the Euclidean distance, $d$, is calculated for each pixel in the computed saliency map. Similarly, the connectedness feature, $c$, is calculated by counting the number of salient neighbors. The probability densities $p(s|d)$ and $p(s|c)$ are fitted using salient object segmentation ground truth from images dataset by Achanta et al. \cite{Achanta2009ImageDataset}. Despite the fact that the proposed algorithm has been reported to yield enhanced saliency detection results, we believe there are four fundamental issues that are overlooked. First, the modeling of the probability densities $p(s|d)$ and $p(s|c)$, being supervised and based on ground truth from images dataset, may not be generally applicable to videos. Second, the uncertainty estimation is based on individual frames of saliency map, thus, losing cues about uncertainty in temporal axis. Third, the method does not offer any degrees of freedom in customizing the estimation process to video content, despite the diverse nature of videos in real-world applications. Finally, an indirect evaluation is performed by showing that the uncertainty-based fusion of spatial and temporal saliency maps is enhanced over other fusion methods. Thus, a direct application-independent performance evaluation methodology is missing.

\par To understand the visual attention mechanism, research usually relies on eye-tracking data analysis to form eye fixation maps. Such maps capture the focus of human subjects watching test videos and potentially correlate well with their visual attention. These maps are often used as the ground truth for saliency in learning-based methods, or as feature space in unsupervised methods. Nevertheless, there has been limited research analyzing the structure of these eye-fixation maps separately from visual stimuli. By studying the eye-fixation maps, we expect to better understand the spatial correlation in video scenes, and henceforth to better understand visual attention mechanisms. The authors in~\cite{Sharma2014} analyzed eye-fixation data of images given location and time sequence of human subjects gaze, using spectral decomposition of the correlation matrix constructed based on eye fixation data of different subjects. Their work shows that the first eigenvector is responsible for roughly $21\%$ of the data, and it correlates well with salient locations in the images dataset. In \cite{Borji2013}, the authors found that it is possible to decode the stimulus category by analyzing statistics (location, duration, orientation, and slope histograms) of fixations and saccades. They used a subset of the NUSEF dataset~\cite{Subramanian2010} containing five categories over a total of 409 images.

\par In this paper, we address the problem of quantifying the uncertainty of detected saliency maps for videos. First, we study spatiotemporal eye-fixation data from the public CRCNS dataset and demonstrate that typically there is high-correlation in saliency between a pixel and its direct neighbors. Then, we propose estimating a pixel-wise uncertainty map that reflects our confidence in the computational saliency map by relating a pixel's value to the values of its direct neighbors in a computationally efficient way. The novelty of this method is that it is unsupervised and independent from the dataset used for testing, which makes it more suitable for generalization. Also, the method exploits information from both spatial and temporal domain, thus, it is uniquely suitable for videos. Moreover, the flexibility of the algorithm parameters allows for customization to specific video content. Additionally, we propose a systematic procedure to evaluate uncertainty estimation performance by explicitly computing uncertainty ground truth in terms of a given saliency map and eye fixations of human subjects watching the associated video segment. 
\section{Analysis of the Eye-Fixation Data}
To motivate the proposed uncertainty estimation method, we present in this section analysis of recorded eye-fixation maps provided in CRCNS as performed in our preliminary study in \cite{alshawi16icme}. In particular, the analysis quantifies the predictability of a pixel in the eye-fixation map given the knowledge of its spatial context. By modeling the pixels of eye-fixation maps and the average of their neighborhood as random variables, we infer the correlation between the eye-fixation map pixels and their immediate $3 \times 3 \times 3$-neighborhood by computing entropy of the eye-fixation pixels versus the entropy of the eye-fixation pixels conditioned on the average of their neighbors. Using the basic properties of entropy, if the neighborhood average completely determines the eye-fixation pixel value then the conditional entropy is equal to zero. Otherwise, the conditional entropy can be any value between zero and a maximum equal to the entropy of the eye-fixation map pixels depending on the correlation between the two quantities (i.e., the pixel value and that of its neighbors). Additionally, to verify the statistical significance of this correlation we compute the entropy of the eye-fixation pixel entropy conditioned on uniformly-distributed random variable.

\par
\begin{figure}
\centering
\captionsetup{justification=centering}
\includegraphics[width=0.5\textwidth,trim={2.4cm 2.9cm 2.2cm 4.0cm},clip]{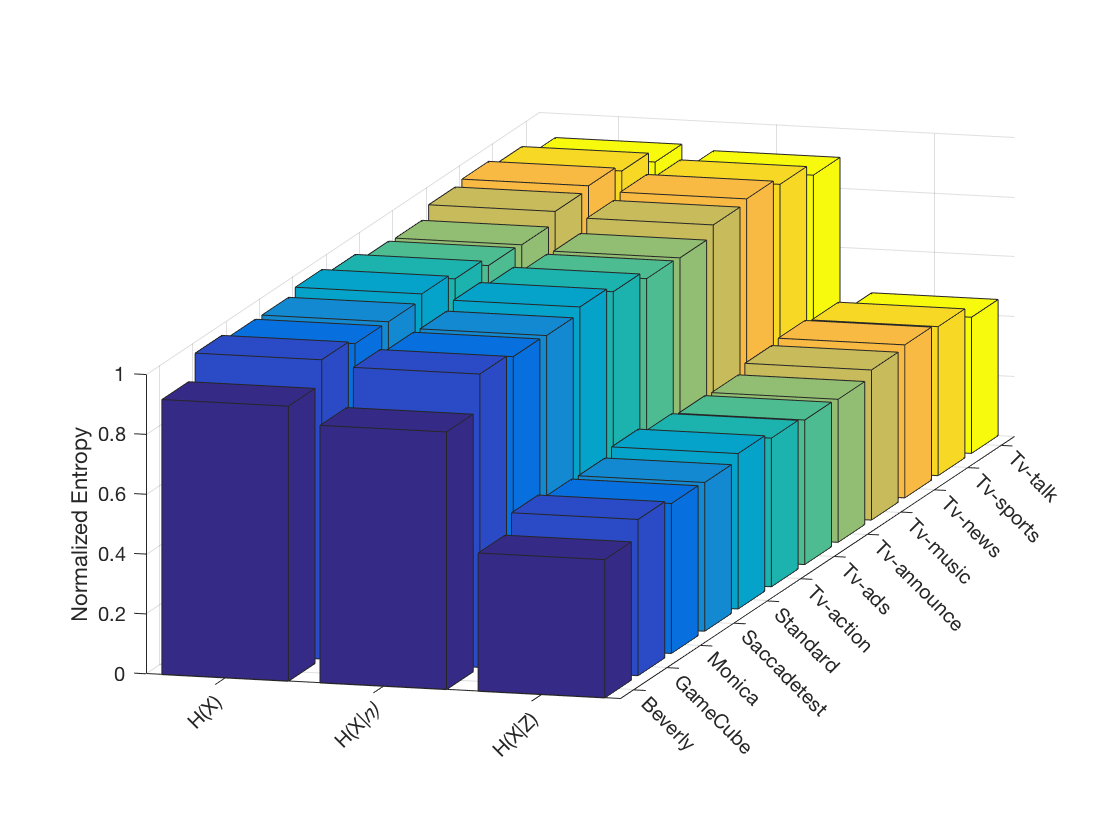}
\caption{Entropy reduction across all videos in CRCNS dataset. Results reported here are computed using \emph{Scale 1} saliency map of size $12 \times 16$.}
\label{fig:EntropyReduction}
\end{figure}

 As reported in \cite{alshawi16icme} and shown in Fig. \ref{fig:EntropyReduction}, in most cases, there is roughly a $50\%$ reduction in entropy when conditioned on neighboring pixels average $H(X|Z)$ compared with the eye-fixation pixels entropy $H(X)$. Notably, the results shown in Fig. \ref{fig:EntropyReduction} are normalized to the highest conditional entropy in the dataset. We can observe from Fig. \ref{fig:EntropyReduction} that the entropy reduction is consistent across the dataset regardless of the video content. Also, to avoid confusing this reduction with a computational limitation error, we compute the entropy of the eye-fixation map given the value of a uniformly distributed random variable $H(X|n)$. As seen in Fig.\ref{fig:EntropyReduction}, the gap between the two conditioned entropy is quite significant indicating the existence of a structure in the eye-fixation maps that can be exploited. Entropy reduction is consistent regardless of probability distribution skewness, as well. This can be shown by redistributing a portion of the probability density from the zero symbol, which dominates eye-fixation maps and explains its low entropy value, to the rest of probability set. This correlation between a pixel value and the average of its neighborhood can be exploited to obtain a rough estimate of a pixel uncertainty in the saliency map given its direct neighborhood, which we detail in the next section.

\section{Unsupervised Uncertainty Estimation using local spatiotemporal neighborhood Cues}
\label{subs:UncertaintyST}
As discussed in the previous section, pixels in eye-fixation maps are correlated and such dependency can be exploited to identify unlikely occurrences in the computational saliency maps. Basically, we assume that visual saliency is consistent and changes in saliency values happen gradually. Thus, sudden changes in saliency value should lower our trust in that particular spatiotemporal event. Thus, saliency map pixels that are significantly different from their neighborhood are most likely uncertain and should be examined more carefully. 
\par However, the size of local neighborhoods crucially depends on the video content. For example, fast action videos would most likely have a small group of contiguous correlated pixels, in the saliency map, around location of the main scene actor. In contrast, a slow changing scene gives the viewers more freedom to explore different parts of the video frame, thus, the corresponding eye fixation map would have a larger group of pixels that are correlated. Therefore, it is important to include uncertainty cues from the appropriate scales in order to more reliably capture context-based events. 
\par In most video saliency detection algorithms, the processing of video frames usually consumes significant computation time. Hence, a common practice is to resize the input video frames to several sizes and define saliency maps generated in terms of the frame scale. It is worth noting that saliency maps generated from size-reduced video frames differ from saliency maps downsampled from saliency maps of higher scale. In the first case, video details lost in the downsampling process are not included in the downsampled saliency map, while in the second case, downsampled saliency maps still maintain such details. Generally, uncertainty estimation should take advantage of saliency maps of multiple scales to enhance the estimation performance. One way to approximate the contribution to uncertainty estimation from different scales is to generate a multi-scale uncertainty map that is a weighted combination of uncertainty generated from different scales. In this paper, we focus our study on how to estimate uncertainty from a single scale.
\par Formally, given a saliency map $\boldsymbol {S}^{(d)}$ of scale $d$ and size $M \times N$ and of depth $K$ frames, we seek to estimate an uncertainty map $\boldsymbol {U}^{(d)}$ of the same scale, size and depth as $\boldsymbol {S}^{(d)}$ that is roughly approximated by saliency value divergence from spatiotemporal local neighborhood mean. The estimation is efficiently computed by processing the map $\boldsymbol {S}^{(d)}$ according to Eq.(\ref{eqn:Spatiotemporal}):

\begin{equation}
\label{eqn:Spatiotemporal}
  \boldsymbol {U}^{(d)} = \gamma\big|\alpha \boldsymbol{S}^{(d)} \ast W^{L_{1} \times L_{2} \times L_{3}}\big|,
\end{equation}
where $\big|.\big|$ is the operation to find the absolute value and $d = 1,2, ... D$ is the scale label, $L_{1} \times L_{2} \times L_{3}$ is the size of the spatiotemporal kernel $W^{L_{1} \times L_{2} \times L_{3}}$, $\alpha$ is a scaling factor for the saliency map to fix its range to be [0,1], and $\gamma$ is a scaling factor for the uncertainty map to ensure the output range is [0,1]. In this paper, we use a simple averaging kernel defined as follows 

\begin{equation}
\label{eqn:STwindow}
  W^{L_{1} \times L_{2} \times L_{3}} =
  \begin{cases} 
      \frac{R-1}{R} & \text{at the center}  \\
      -\frac{1}{R}, & otherwise, 
   \end{cases}
\end{equation}
where $R$ = $L_{1} \times L_{2} \times L_{3}$. The design of $W^{L_{1} \times L_{2} \times L_{3}}$ can be viewed as the difference between saliency value and a moving average window of size $L_{1} \times L_{2} \times L_{3}$. $W^{L_{1} \times L_{2} \times L_{3}}$ , with appropriate size, can follow the changes in the scene and, to some extent, approximates the common trend of pixel saliency change over time.

In order to systematically analyze spatiotemporal uncertainty estimation, we study the contribution of spatial neighbors separate from temporal neighbors which might lead to a better understanding of spatial context in saliency maps. Thus, we introduce in the following subsections two special cases of the proposed algorithm: uncertainty estimation from temporal cues and uncertainty estimation from spatial cues. Relying only on temporal neighbors, the proposed algorithm estimates the uncertainty of a pixel in frame $k$ by studying its correlation with its neighbors in the same location across all $K$ frames, as we have proposed in \cite{alshawi15temporal}. By dividing the saliency map into temporal neighborhoods, we can treat each pixel location as separate 1-D signal that can be processed using a simple 1-D filter of length $L_t$ to calculate pixel-neighborhood divergence. Similarly, we can divide the saliency map into spatial neighborhoods that span $L_{s_1} \times L_{s_2}$ pixels in a single frame, as we have reported in \cite{alshawi15spatial}.

\subsection{Uncertainty Estimation from Temporal Cues}
\label{subs:UncertaintyT}
 For a given saliency map $\boldsymbol S$ of size $M \times N$ and of depth $K$ frames, we decompse the map into 1-D signals as follows

\begin{equation}
\label{eqn:Stemporal}
\boldsymbol S =
\begin{bmatrix}
    s[1,1] & s[1,2]& \dots & s[1,n]& \dots  & s[1,N]\\
    s[2,1] & s[2,2]& \dots & s[2,n]& \dots  & s[2,N] \\
    \vdots &\vdots & \dots & \vdots& \dots & \vdots \\
    s[m,1] & s[m,2]& \dots & s[m,n]& \dots  & s[2,N] \\
    \vdots &\vdots & \dots & \vdots& \dots & \vdots \\
    s[M,1]& s[M,2] &\dots  & s[M,n]&\dots  & s[M,N]
\end{bmatrix},
\end{equation}
where $m$ = $1,2,...,M$, $n$ = $1,2,...,N$, are the spatial coordinates of the saliency map.
\par We seek to construct an uncertainty map $\boldsymbol U$ of the same size and depth as $\boldsymbol S$ by iteratively processing 1-D signals $\boldsymbol s$ located at saliency map pixel $[m,n]$  according to 
\begin{equation}
\label{eqn:MA}
U[m,n] = \gamma\big|\alpha S[m,n] \ast W^{L_t}\big|,
\end{equation}
where $m$ = $1,2,...,M$, $n$ = $1,2,...,N$, are the spatial coordinates of both the saliency map and uncertainty map, $\alpha$ and $\gamma$ are scaling factors, and $W^{L_t}$ is the temporal filter of length $L_t$, defined by
\begin{equation}
\label{eqn:windowT}
W^{L_t} = [\frac{-1}{L_t} ... \frac{-1}{L_t}, \frac{L_t-1}{L_t}, \frac{-1}{L_t} ... \frac{-1}{L_t}],
\end{equation}

\subsection{Uncertainty Estimation from Spatial Cues}
\label{subs:UncertaintyS}
Similar to the temporal neighborhood case, given a saliency map $\boldsymbol S$ (Eq. (\ref{eqn:Sspatial})) of size $M \times N$ and of depth $K$ frames, we construct an uncertainty map $\boldsymbol U$ (Eq. (\ref{eqn:Uspatial})) of the same size and depth as $\boldsymbol S$ by iteratively processing saliency frames $S_{k}$ using a 2-D averaging kernel $W^{L_{s_1} \times L_{s_2}}$ (Eq.(\ref{eqn:Kernel})) of size $L_{s_1} \times L_{s_2}$. 

\begin{equation}
\label{eqn:Sspatial}
\boldsymbol S =
\begin{bmatrix}
    S_{1} & S_{2} & \dots & S_{K}
\end{bmatrix},
\end{equation}
\begin{equation}
\label{eqn:Uspatial}
\boldsymbol U =
\begin{bmatrix}
    U_{1} & U_{2} & \dots & U_{K}
\end{bmatrix},
\end{equation}
\begin{equation}
\label{eqn:Kernel}
    U_{k} = \gamma\big|\alpha S_{k} \ast W^{L_{s_1} \times L_{s_2}}\big|,
\end{equation}
where $k$ = $1,2,...,K$ is the frame index, $W^{L_{s_1} \times L_{s_2}}$ is a spatial filter similar to averaging kernel $W^{L_t}$, symmetrical around its center and has a size of $L_{s_1} \times L_{s_2}$, $\alpha$ and $\gamma$ are scaling factors.

\section{Methods for Ground Truth Generation and Performance Evaluation}
To objectively evaluate the performance of an uncertainty estimation algorithm, ideally we need to compare the estimated uncertainty against the ground truth, or the true uncertainty. However, such true uncertainty data is not readily available.

\subsubsection{Computing True Uncertainty}
\begin{figure}
\begin{center}
\begin{tabular}{c c c}
\includegraphics[height=2.7cm]{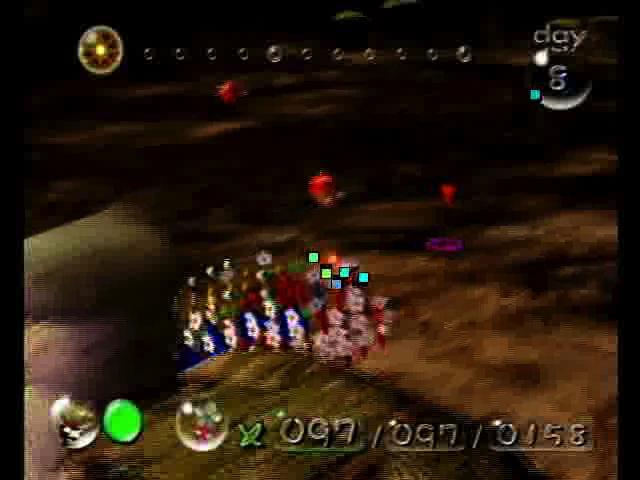} & \includegraphics[height=2.7cm]{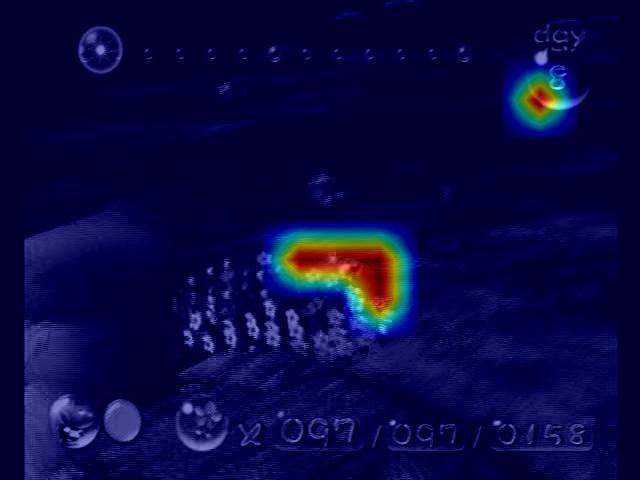} & \includegraphics[height=2.7cm,trim={9.5cm 7.5cm 10cm 7.5cm},clip]{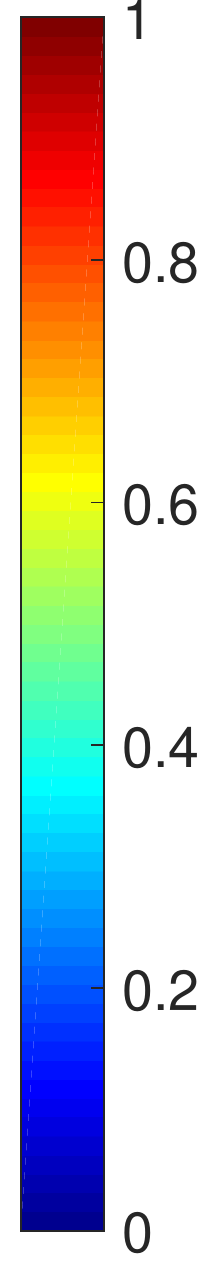}\\
(a) & (b) &\\
\includegraphics[height=2.7cm]{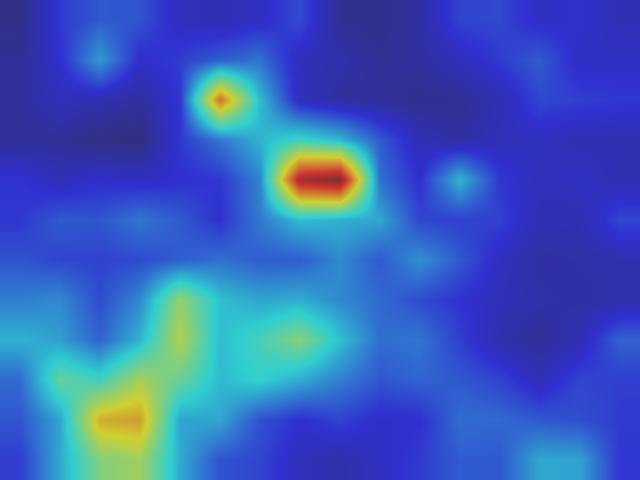} & \includegraphics[height=2.7cm]{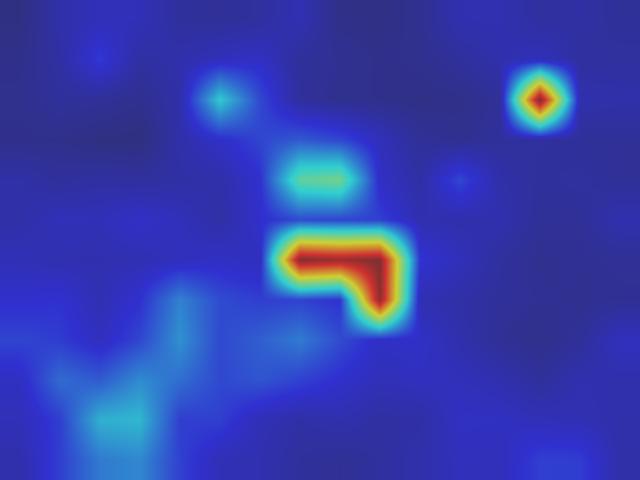} & \includegraphics[height=2.7cm,trim={9.5cm 7.5cm 10cm 7.5cm},clip]{Figures/colorbar7.pdf}\\
(c) & (d) &\\
\end{tabular}
\end{center}
\caption
{\label{fig:trueUncertainty}
Examples illustrating true uncertainty data. (a) Original video frame with eye fixation superimposed (small color squares in the center and top-right corner); (b) Resized eye fixation map superimposed on the original frame; (c) Saliency detection results; (d) True uncertainty. We note that the color display is only for a better illustration, which involves some interpolation causing the discrete resized fixation map to appear continuous.}
\end{figure}

\begin{figure}
\centering
\includegraphics[height=7cm]{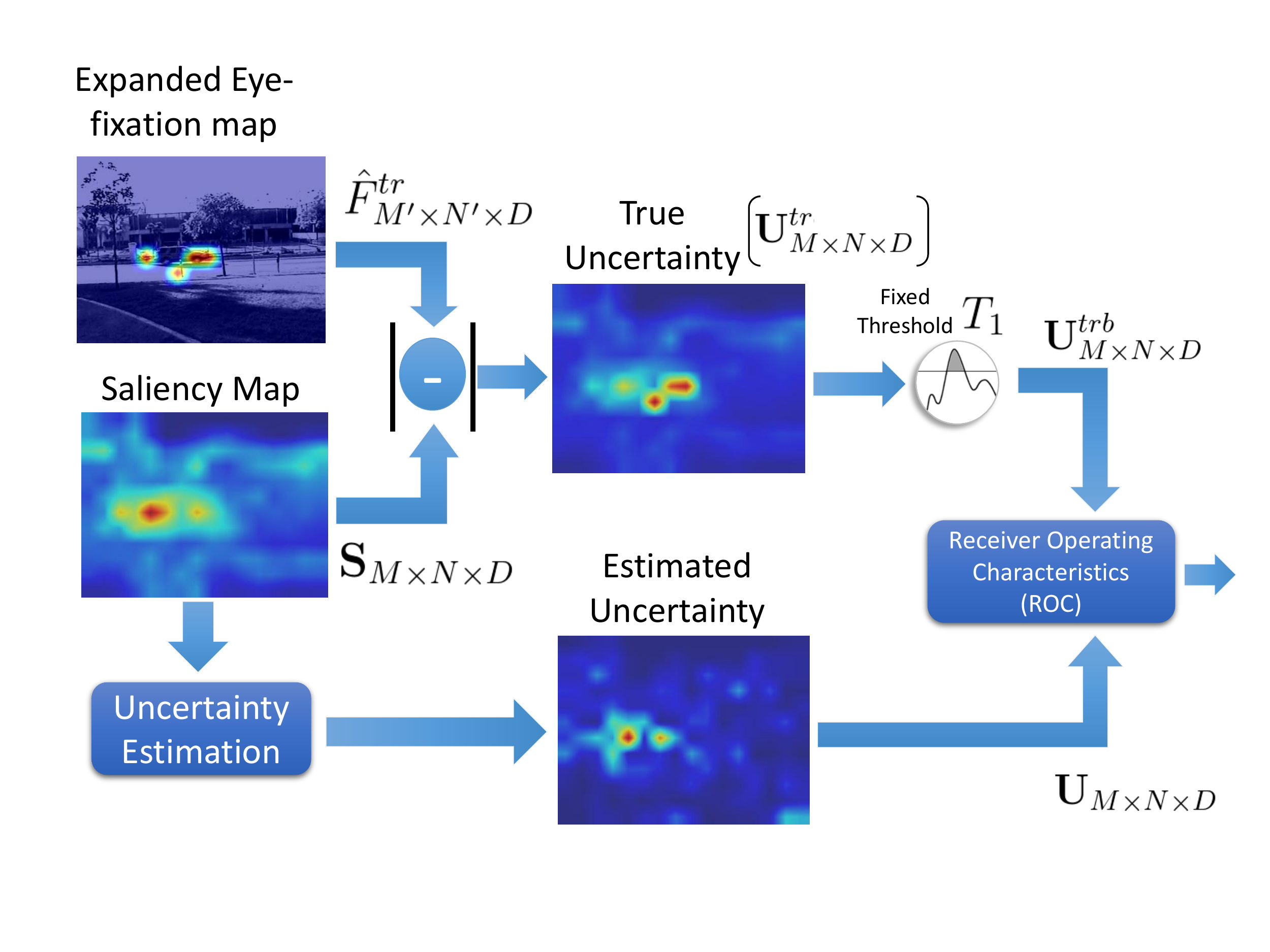}
\caption{Evaluation methodology \cite{alshawi15temporal}.}
\label{fig:evaluation}
\end{figure}

Available databases for saliency detection research usually contain ground truth data recording eye fixations of human subjects viewing the images or videos. Based on the eye fixation data, as we proposed in \cite{alshawi15temporal}, the following method is used to generate the true uncertainty data. Fig. \ref{fig:trueUncertainty} illustrates this procedure with some examples while the block diagram is shown in Fig.\ref{fig:evaluation}. 
First, we compile the fixation data from all subjects in CRCNS dataset into a single map $\boldsymbol{\hat{F}}^{tr}$ of size $M'$, $N'$, and $K$ being the height, width, and the total number of frames, respectively. We add $1$ to $\hat{F}^{tr}[i,j,k]$ for every eye fixation that corresponds to pixel location $[i,j,k]$. Second, we resize the fixation map $\boldsymbol{\hat{F}}^{tr}$ to $M$, $N$ and $K$; the respective height, width, and depth of the saliency map $\boldsymbol{S}$ from a saliency detection algorithm. This resizing is necessary because many saliency detection techniques work on downsampled video frames for computational efficiency. However, for the binary map $\boldsymbol{\hat{F}}^{tr}$, the resizing is not exactly a downsampling procedure.Denoted as $\boldsymbol{F}^{tr}$, the resized binary fixation map is obtained as follows
\begin{equation}
F^{tr}[m,n,k]=\sum_{\forall(i,j) \in \Phi[m,n,k]} \hat{F}^{tr}[i,j,k],
\end{equation}
where $\Phi[m,n,k]$ is an indexing function that points to the set of pixels in $\hat{F}^{tr}$ that corresponds to pixel $[m,n,k]$ in $F^{tr}$ map. Here, we use the sum of eye-fixation points from all subjects so that salient locations agreed upon by majority of subjects have the highest saliency, but at the same time sparse ``1"s in the original fixation truth data are not lost. Finally, assuming that the saliency map $\boldsymbol S$ is normalized, we normalize $\boldsymbol{F}^{tr}$ and calculate the true uncertainty as
\begin{equation}
\label{eqn:trueUncert}
\boldsymbol{U}^{tr}=\big|\boldsymbol{S}-\boldsymbol{F}^{tr}\big|.
\end{equation}

Obviously, $\boldsymbol{U}^{tr}$ shows how far each saliency estimate is from the recorded fixations. Thus, it can serve as a measure of the estimation uncertainty. Even though the individual eye-fixation data is binary, the aggregated fixation maps $\boldsymbol{\hat{F}}^{tr}$, $\boldsymbol{F}^{tr}$, the derived true uncertainty data $\boldsymbol{U}^{tr}$, and the saliency detection results $\boldsymbol{S}$ are continuous values.

\subsubsection{Performance Measurement}
\par With the true uncertainty data available, we use a detection theory-based scheme for the performance evaluation \cite{alshawi15temporal}. The scheme generates an ROC curve and uses AUC as the performance metric \cite{hastie2005elements}. Since our true uncertainty data $\boldsymbol{U}^{tr}$ is continuous, it needs to be converted to binary data, denoted as $\boldsymbol{U}^{trb}$, as the ROC curve is intended for binary classifiers. This conversion is conducted by applying a threshold $T_1$. To generate the ROC curve, the uncertainty estimates $\boldsymbol{U}$ are also thresholded by $T_2$ into a binary form, $\boldsymbol{U}^b$, and compared against $\boldsymbol{U}^{trb}$. Thus, both the true detection rate (TDR) and the false positive rate (FPR) are obtained. When we change the value of $T_2$, sweeping through its whole range, pairs of TDR and FPR are obtained to yield an ROC curve plotted as TDR vs. FPR. Then, the AUC is easily computed. AUC ranges between $0$ and $1$, with a greater value indicating better performance, and $0.5$ indicating a performance equivalent to random classifier.

\section{Experiments}
\label{sec:exp}
 We conducted three sets of experiments to study several aspects of the proposed algorithm. In the first set, we compare the relative performance based on the neighborhood selection. We evaluate and compare the performance of the proposed algorithm using: 
 \begin{itemize}
  \item Spatiotemporal neighborhood as described in \ref{subs:UncertaintyST}, labeled Spatiotemporal Uncertainty (\emph{STU})
  \item Temporal neighborhood as described in \ref{subs:UncertaintyT}, labeled Temporal Uncertainty (\emph{TU}) \cite{alshawi15temporal}
  \item Spatial neighborhood as described in \ref{subs:UncertaintyS}, labeled Spatial Uncertainty (\emph{SU}) \cite{alshawi15spatial}
  \item Naive fusion of Spatial and Temporal Uncertainty (\emph{SU+TU}), a pixel-wise addition of \emph{TU} and \emph{SU} maps
  \item Entropy-based Uncertainty (\emph{EU}) \cite{Fang2014video}
  \item Local variance of spatiotemporal neighborhood, labeled \emph{Baseline}
\end{itemize}
The performance of these algorithms is quantified in terms of Area-Under-the-Curve (AUC) values of their corresponding Receiver-Operating-Characteristic (ROC) curves. We, also, show effects of saliency map scale as well as kernel size on the proposed algorithm's performance. Details on data and experiments procedure are provided in the dataset section and the performance evaluation methodology section, respectively. The second set of experiments are designed to show performance of the proposed uncertainty estimation algorithm given different categories of videos. Also, we show the distinct effects of kernel size on the proposed algorithm performance given radically different video contents. The third set of experiments verifies the performance of the proposed algorithms using additional datasets and saliency detection models.

\subsection{Datasets}
 We tested the proposed unsupervised uncertainty estimation algorithm using three publicly available databases: CRCNS \cite{CRCNS05}, DIEM \cite{MitalCogComp2011}, and AVD \cite{MarighettoICIP2017}. The CRCNS \cite{CRCNS05} database includes 50 videos, with the resolution being $480 \times 640$ and the duration ranging from $5$ to $90$ seconds with $30$ frames per second. The videos contents are diverse with a total of $12$ categories ranging from street scenes to video games and from TV sports to TV news. In many cases the videos contain variations of lighting conditions, severe camera movements, and high motion blur effects. Eye fixation data are provided with each video, recorded for a group of eight human subjects watching the videos under task-free view condition. The DIEM \cite{MitalCogComp2011} database includes 85 videos, with varying resolutions and duration up to $130$ seconds with $30$ frames per second. The videos content are mainly limited to TV and film content including film trailers, music videos, and advertisement. The eye fixation data are collected from 250 participants under task-free view conditions. The AVD \cite{MarighettoICIP2017} database includes 148 videos, with varying resolutions and mean duration of $22$ seconds with $30$ frames per second. The video contents are limited to moving objects, landscape, and faces. The eye fixations data are collected from 176 observers. The AVD dataset contains two sets of videos of the same visual content but one with audio and the other without. According to their findings on the effect of audio on the attention of the participants, we only select the videos without associated audio.
 
 For our experiments, we generated saliency maps for the videos using a recent algorithm based on 3D FFT local spectra (3DFFT) \cite{long2015saliency}. However, for validation, we also share the results from two additional saliency models: STSR \cite{Seo2009JoV} and PQFT \cite{Guo2008CVPR}, which are shown at the end of this section. Unless stated otherwise, saliency maps used in all experiments are generated using 3DFFT.  In most of our experiments, the saliency maps are reduced in size to three different scales. \emph{Scale 1} is of size $12 \times 16$; a downscale of frames original size $480 \times 640$, where every $40 \times 40$ region in the original frame corresponds to a single pixel in \emph{Scale 1}. Similarly, \emph{Scale 2} saliency maps are $24 \times 32$, where every pixel is equivalent to $20 \times 20$ region of pixels in the original sized frame, and \emph{Scale 3} saliency maps are $48 \times 64$, where every pixel is equivalent to $10 \times 10$ regions.

\subsection{Results and Discussions}
\par
\begin{figure}
\begin{center}
\includegraphics[height=0.38\textwidth]{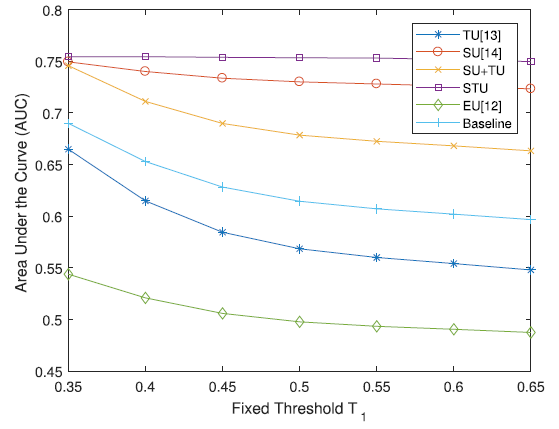} 
\end{center}
\caption
{\label{fig:threshold_fig}
Examples illustrating that relative uncertainty estimation performance is independent of fixed threshold $T_1$ applied to true uncertainty. Results reported here were generated using \emph{Scale 1} maps with averaging kernel of length $5$ for TU, of size  $5 \times 5$ for SU, and $5 \times 5 \times 5$ for STU.}
\end{figure}

 The performance evaluation procedure described earlier utilizes a fixed threshold $T_1$ to transform the continuous valued true uncertainty $\boldsymbol{U}^{tr}$ to binary ground truth. First, we examine the impact of changing the value of $T_1$. The algorithms under consideration are: Temporal Uncertainty (\emph{TU}), Spatial Uncertainty (\emph{SU}), Fused Spatial and Temporal Uncertainty (\emph{SU+TU}), Spatiotemporal Uncertainty (\emph{STU}), Spatiotemporal local variance (\emph{Baseline}) computed on the same neighborhood as STU, and Entropy-based Uncertainty (\emph{EU}). Fig.\ref{fig:threshold_fig} shows the performance of these algorithms in terms of AUC versus $T_1$. 
As shown in Fig.\ref{fig:threshold_fig}, $T_1$ directly affects AUC value; as the value of $T_1$ increases, the AUC value of all algorithms considered here decreases. It is also interesting to point out that the gradient of AUC levels-off as $T_1$ reaches higher values. Although we can see that $T_1$ value significantly changes AUC, conclusions based on relative AUC values are consistent regardless of the value of $T_1$. As shown in Fig.\ref{fig:threshold_fig}, STU outperforms all other algorithms while EU is performing the worst in this experiment. Please note that the reported AUC results are for \emph{Scale 1} maps with averaging kernel of length $5$ for TU, of size  $5 \times 5$ for SU, and $5 \times 5 \times 5$ for STU.

\subsubsection{Neighborhood Selection (domain,scale,size)}
\label{subss:neighborhoodSelection}
\par
\begin{figure}
\begin{center}
\includegraphics[width=0.4\textwidth,trim={7cm 3cm 9cm 1cm},clip]{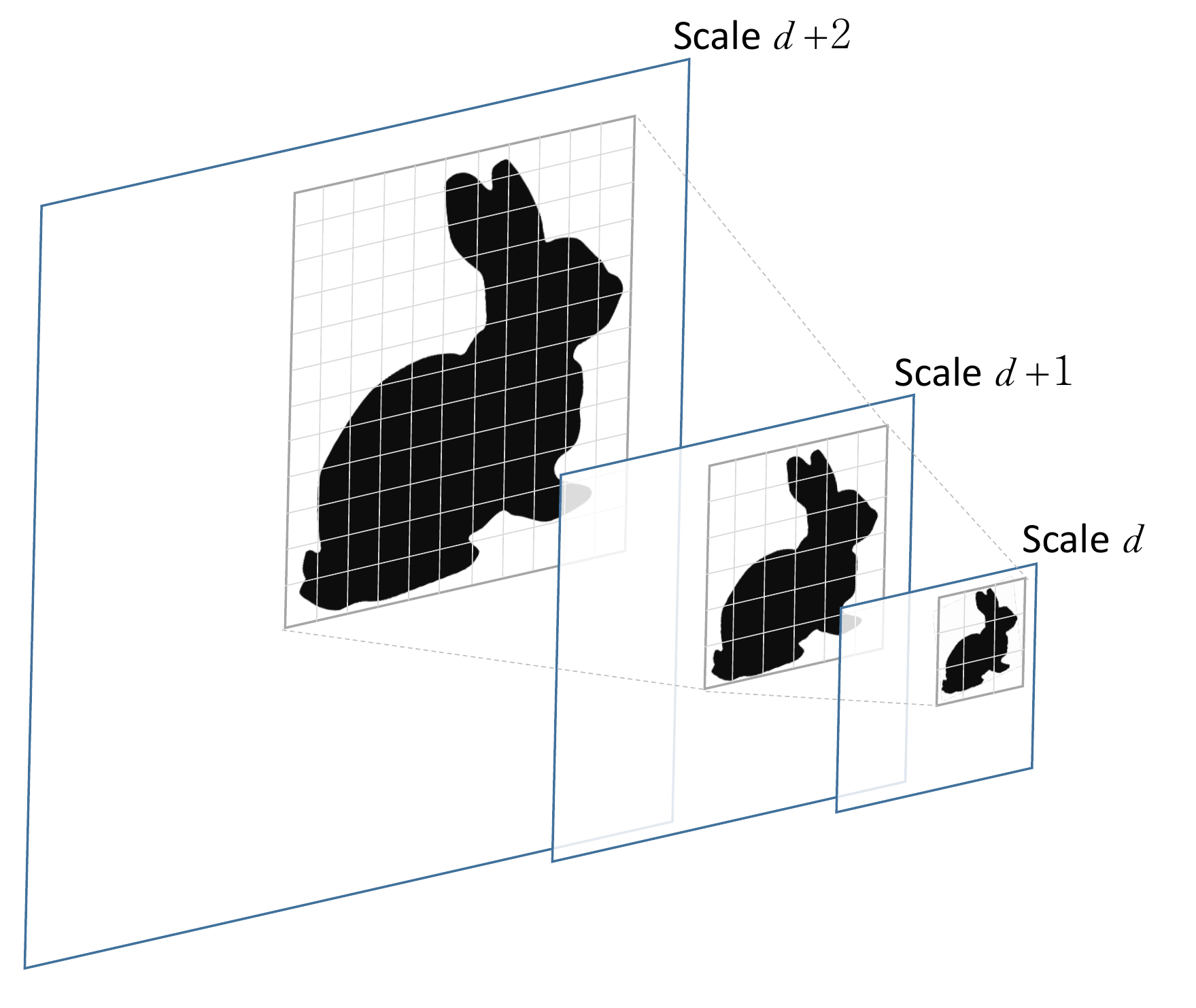}
\end{center}
\caption
{\label{fig:SupportRegion}
Kernel size changes between scales according to support region size.}
\end{figure}

\begin{figure}
\begin{center}
\includegraphics[height=0.38\textwidth]{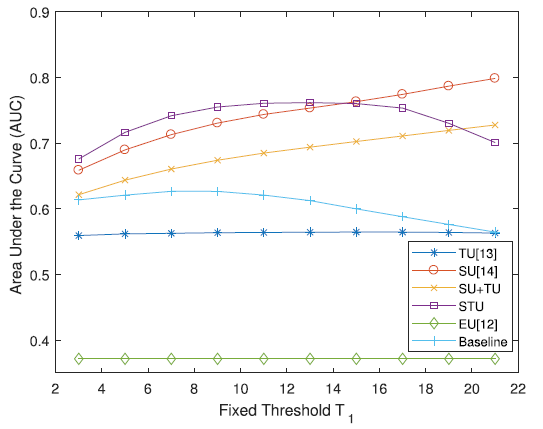}
\end{center}
\caption
{\label{fig:Kernel_Selection_fig}
AUC value is affected by the choice of the kernel size at the same scale. Results reported here use \emph{Scale 2} saliency maps and $T_1$ = $0.55$.}
\end{figure}

\begin{figure}[!htp]
\begin{center}
\includegraphics[height=0.38\textwidth]{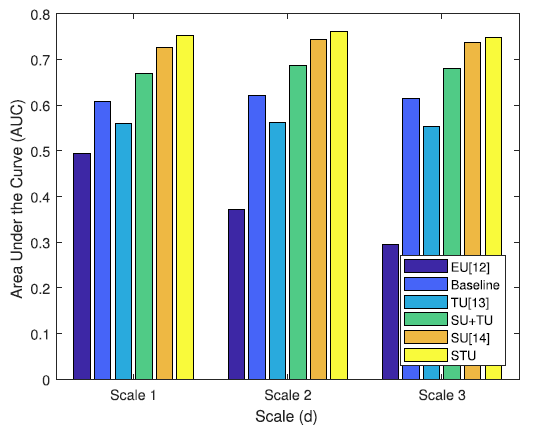}
\end{center}
\caption
{\label{fig:scale_Selection_fig}
The impact of scale change with constant support region (using different kernel sizes). AUC value is relatively the same when processing different scales. Results reported here use threshold $T_1$ = $0.55$.}
\end{figure}

 As shown earlier, in addition to the threshold $T_1$, the neighborhood selection affects AUC value. Additionally, scale of the saliency maps and size of the processing kernels affect the performance of proposed estimation algorithm as well. In Fig. \ref{fig:scale_Selection_fig}, we show the AUC values for the algorithms under test using different saliency map scales. The experiment is conducted using saliency maps of \emph{scale 1}, \emph{2} and \emph{3} and an averaging kernel. In order to fix the kernel size relative to the support region size in the original frame, we use different kernel size for each scale, as illustrated in Fig. \ref{fig:SupportRegion}. In Fig. \ref{fig:scale_Selection_fig}, \emph{Scale 1} experiment uses $5 \times 5$ for SU and $5 \times 5\times 5$ for STU. Similarly, for \emph{Scale 2}: $11 \times 11$ for SU and $11 \times 11\times 5$ for STU, and for \emph{Scale 3}: $21 \times 21$ for SU and $21 \times 21\times 5$ for STU. The length of TU kernel is fixed $L_t$ = $5$. We can see that the change in AUC value is relatively small, thus, shows the effectiveness of the proposed uncertainty algorithm even when saliency maps are considerably small size. This feature of the proposed estimation algorithm can be exploited to reduce the required computations, thus speeding up the estimation process without much sacrifice in terms of performance. Please note that AUC value for EU algorithm changes over different scales, due to true uncertainty $\boldsymbol{U}^{tr}$ containing more details as the scale increases.
Moreover, kernel size affects the performance of the proposed algorithm as well. Fig. \ref{fig:Kernel_Selection_fig} shows the performance of the estimation algorithms under test, in terms of AUC values, when the estimation kernel size is changed. The experiment is conducted using \emph{scale 2} saliency map and variable kernel size $r$ ($r$ for \emph{TU}, $r \times r$ for \emph{SU}, and $r \times r \times r$ for \emph{STU}). As shown in Fig. \ref{fig:Kernel_Selection_fig}, AUC of the proposed algorithm changes as the size of the kernel changes. However, the change in \emph{TU} performance is significantly smaller than that of \emph{SU} and \emph{STU} because the number of pixels added into \emph{SU} and \emph{STU} kernels is significantly more than the number of pixels added to \emph{TU} kernel. There is, however, a slight degradation in \emph{TU} performance as the kernel size increases (starting from $L_t$ = $13$ onwards), which can be attributed to including less relevant pixel in the estimation process as the kernel size increase. For kernels of sizes $3 \times 3 \times 3$ till $11 \times 11 \times 11$, it can be seen that \emph{STU} achieves higher AUC than \emph{SU}. However, such trend inverts starting from kernel size $13 \times 13 \times 13$ onwards. This could be explained by noting the similar trend in \emph{TU} as the kernel size increases in time domain due to inclusion of pixels that might be less relevant. The performance degradation in \emph{STU} (and \emph{Baseline} as well) is more profound than \emph{TU} because, for a kernel size of $n \times n \times n$, $n^2$ pixels are added to \emph{STU} estimation process for every additional frame while only a single pixel is added for \emph{TU} estimation. It is important here to clarify that these results are obtained for the whole dataset (50 videos). Thus, trends that are observed here are not necessarily true for every video type. We discuss in details the performance as related to the video categories in the next section.

\subsubsection{Video Categories (domain,size)}
\label{subss:videoCat}
Given the diverse nature of scenes and dynamics in the dataset, we  evaluate the performance of our proposed algorithm for each category in the dataset. For these experiments, we set $T_1$ = $0.55$ and use \emph{Scale 1} saliency maps. Table \ref{table:categoryPerformance} shows AUC values for \emph{TU} ($L_t$ = $5$), \emph{SU} ($L_{s_x}$ = $5$), \emph{ST+SU}, \emph{STU} ($L_{st_x}$ = $5$), \emph{EU}, and \emph{Baseline} ($L_{st_x}$ = $5$), for each category, separately. As shown in Table \ref{table:categoryPerformance}, AUC values for the proposed algorithm are above $0.5$, indicating that the proposed algorithm is advantageous over random guessing. Additionally, the algorithm performs better than \emph{EU} in every category and in some by a wide margin. One interesting result is that AUC for \textsf{Saccadetest} video is significantly higher than other categories for all algorithms considered here. This can be attributed to its non-complex structure, which shows a disk moving against a light textured background. Notably, \emph{STU} achieves highest performance in every category except \textsf{Saccadetest}. This could be attributed to its relative constant scenes in the first segment of the video.
\par
\begin{figure}[!htp]
\begin{center}
\includegraphics[height=0.38\textwidth]{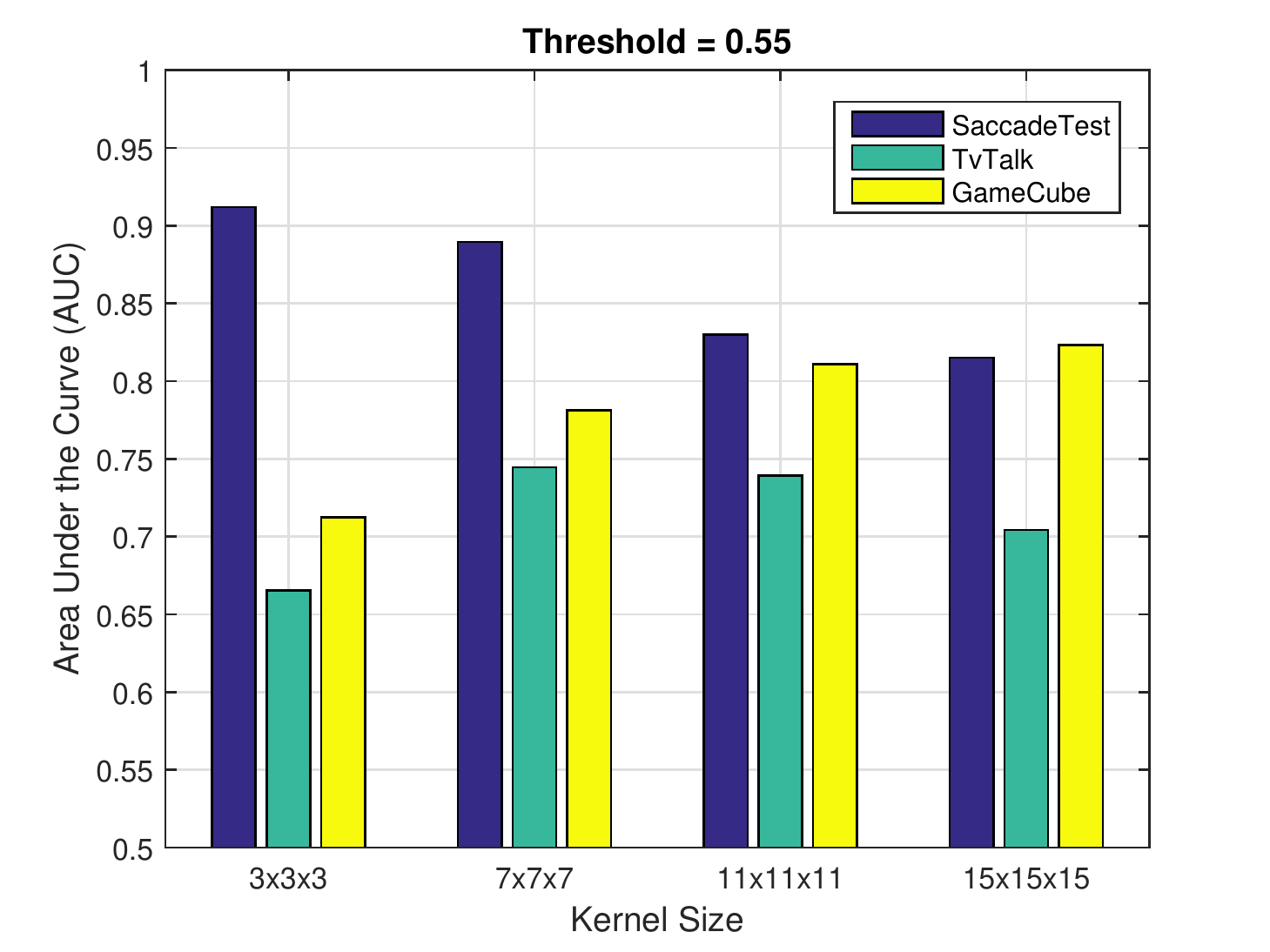}
\end{center}
\caption
{\label{fig:catVsize_fig}
Examples illustrating the effect of kernel size on the estimation performance using \emph{STU} extracted from uncertainty maps of \emph{Scale 2}.}
\end{figure}

 Moreover, we explore the effect of kernel size on the estimation performance. In these experiments, we focus on \emph{STU}, however, \emph{TU}, \emph{SU}, and \emph{SU+TU} exhibit similar behavior. Fig.\ref{fig:catVsize_fig} shows AUC for \emph{STU} estimation algorithm on three video categories; \textsf{saccadetest}, \textsf{tv-talk}, and \textsf{gamecube} for kernel sizes: $L_{st_x}$ = $3$, $7$, $11$, and $15$, using \emph{Scale 2} saliency maps and $T_1$ = $0.55$. As shown in Fig.\ref{fig:catVsize_fig}, as the kernel size increases, \emph{STU} performance on \textsf{saccadetest} degrades indicating that the relevance saliency context in \textsf{saccadetest} video is strictly local and including more pixels than direct neighbors degrades uncertainty estimation performance. Indeed, the structure of \textsf{saccadetest} video justifies these results due to its simplicity. In contrast, \textsf{gamecube} video uncertainty estimation results increase as the kernel size increase. This indicates that the set of correlated saliency pixels for \textsf{gamecube} is larger than its direct neighbors. The large set of correlated saliency pixels in \textsf{gamecube} might be explained by its complex structure and the fact that these videos contain multiple salient actors in the same scene making it more difficult to capture saliency context from small local neighborhoods. On the other hand, \emph{STU} performance in estimating uncertainty for \textsf{tv-talk} reaches maximum level in intermediate kernel sizes and then decreases as we increase the kernel size, indicating that the most appropriate kernel size to capture relevant saliency context is half the frame size.

\subsubsection{Comparison across various datasets and saliency models}
\label{subss:comparisonDatasets}
\par
\begin{figure}[!htp]
\begin{center}
\includegraphics[height=0.38\textwidth]{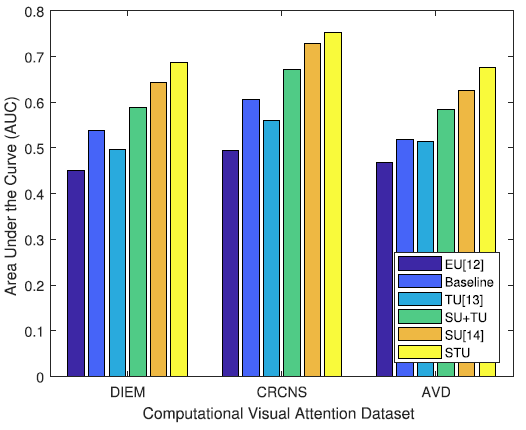}
\end{center}
\caption
{\label{fig:datasets}
The performance of the proposed algorithm across the datasets CRCNS \cite{CRCNS05}, DIEM \cite{MitalCogComp2011}, and AVD \cite{MarighettoICIP2017}. }
\end{figure}

 In this section, we present evaluation results for the proposed algorithm across various datasets. We compare the performance of the proposed algorithm using videos from three datasets: CRCNS \cite{CRCNS05}, DIEM \cite{MitalCogComp2011}, and AVD \cite{MarighettoICIP2017}. Fig. \ref{fig:datasets} shows the the AUC values of the five uncertainty estimation methods using videos from the three datasets. In Fig. \ref{fig:datasets}, \emph{STU} performance is the highest among all datasets. In general, the trend and ranking between the uncertainty estimation methods is consistent across the three datasets. 
\par
\begin{figure}[!htp]
\begin{center}
\includegraphics[height=0.38\textwidth]{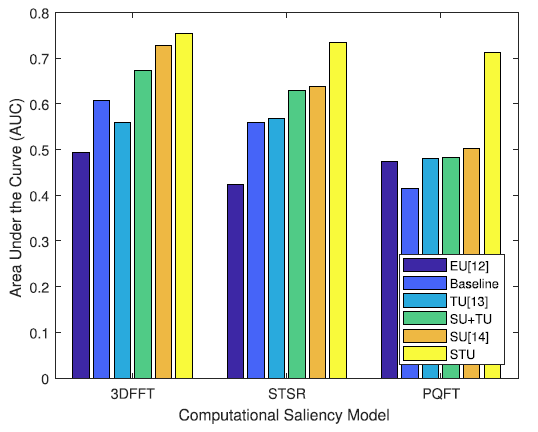}
\end{center}
\caption
{\label{fig:models}
The performance of the proposed algorithm using the saliency models 3DFFT \cite{long2015saliency}, STSR \cite{Seo2009JoV}, PQFT \cite{Guo2008CVPR}}
\end{figure}

 Additionally, we present the evaluation results for the proposed algorithm across using three saliency models: 3DFFT \cite{long2015saliency}, STSR \cite{Seo2009JoV}, and PQFT \cite{Guo2008CVPR}. Fig. \ref{fig:models} shows the AUC values of the five uncertainty estimation algorithms. In Fig. \ref{fig:models}, a consistent trend and ranking between the five algorithms exist across all three saliency models, where \emph{STU} achieves the highest AUC value.
\par Moreover, we evaluate the proposed algorithm, in terms of the computed uncertainty map distribution versus uncertainty ground truth maps distribution, using four distribution-based metrics; Jeffrey Divergence (JD), Jensen-Shannon divergence (JS), Histogram Intersection (HI), L2-norm. As shown in Table.\ref{table:distanceMetrics}, the proposed algorithm provides the closest distribution to that of the ground truth maps across all four metrics and all datasets. 

\section{Conclusion}
In this paper, we discussed the problem of quantifying uncertainty for video saliency detection. To solve this problem, we presented an algorithm to estimate pixel-wise uncertainty in computational saliency maps, which relies on a common feature of human fixation. Our experiments, using CRCNS dataset, showed a reduction of roughly $50\%$, across all videos, in pixel entropy when conditioned on its local neighbors' average. The experiment shows that local correlation exists in saliency perceived by HVS. Thus, saliency map's pixels ought to be highly correlated with their local neighbors. Using this result, we formulated the proposed algorithm according to temporal, spatial, and spatiotemporal neighborhoods and studied the effect of neighborhood selection on the algorithm performance. Additionally, we showed that the appropriate size of local neighborhood is mainly determined by the video content and makes a significant impact on the algorithm performance. For performance evaluation, we proposed a systematic performance evaluation scheme including the generation of true uncertainty and ROC curve-based objective assessment. The proposed algorithm outperforms state-of-the-art uncertainty estimation algorithms across three different datasets: CRCNS, DIEM, and AVD. Consistent performance has been observed with different saliency models. Our algorithm is unsupervised and computationally highly efficient. Additionally, the performance of proposed algorithm could be further enhanced by using a weight-average combination of uncertainty maps from different scales depending on the video content, which we did not explore in this study. The proposed algorithm can be very useful, either as a stand-alone objective evaluation method for saliency detection algorithms, or as an effective means of quality control for saliency-based video processing applications.

\begin{table*}[h]
\centering
\caption
{\label{table:categoryPerformance}
List of AUC value for different categories using fixed threshold $T_1= 0.55$ and \emph{Scale 1} saliency maps. Note that the highest AUC value in each category is labeled in \textcolor{ForestGreen}{green} and lowest AUC value in \textcolor{red}{red}. Also, the category with the highest AUC in the dataset is shown in \textbf{bold}} 
\begin{tabular}{|c|c|c|c|c|c|c|}
\hline
 & \emph{TU} \cite{alshawi15temporal} & \emph{SU} \cite{alshawi15spatial} & \emph{SU+TU} & \emph{STU} & \emph{EU} \cite{Fang2014video} & \emph{Baseline}\\
 \hline \hline
 \textsf{beverly} 		& \textcolor{red}{0.5793} & 0.8088 & 0.7174 & \textcolor{ForestGreen}{0.8130} & 0.5835 & 0.6915\\ \hline  
 \textsf{gamecube} 		& 0.5987 & 0.7636 & 0.7155 & \textcolor{ForestGreen}{0.7913} & \textcolor{red}{0.5906} &  0.6834\\ \hline
 \textsf{monica} 		& 0.6152 & 0.7801 & 0.7240 & \textcolor{ForestGreen}{0.7994} & \textcolor{red}{0.5728} & 0.6506\\ \hline 
 \textbf{\textsf{saccadetest}} 	& \textbf{\textcolor{red}{0.7722}} & \textbf{\textcolor{ForestGreen}{0.8734}} & \textbf{0.8216} & 0.8587 & \textbf{0.8458} & \textbf{0.8308}\\ \hline
 \textsf{standard} 		& 0.5866 & 0.7190 & 0.6609 & \textcolor{ForestGreen}{0.7462} & \textcolor{red}{0.5165} & 0.5841\\ \hline
 \textsf{tv-action} 	& 0.7481 & 0.8466 & 0.7970 & \textbf{\textcolor{ForestGreen}{0.8667}} & 0.7245 & \textcolor{red}{0.6491}\\ \hline
 \textsf{tv-ads} 		& 0.5565 & 0.7248 & 0.6565 & \textcolor{ForestGreen}{0.7476} & \textcolor{red}{0.5228} & 0.5360\\ \hline
 \textsf{tv-announce} 	& 0.4555 & 0.6679 & 0.5550 & \textcolor{ForestGreen}{0.7321} & \textcolor{red}{0.4434} & 0.5818\\ \hline
 \textsf{tv-music} 		& 0.5548 & 0.6721 & 0.6236 & \textcolor{ForestGreen}{0.7427} & \textcolor{red}{0.4471} & 0.5771\\ \hline
 \textsf{tv-news} 		& 0.5051 & 0.6497 & 0.5885 & \textcolor{ForestGreen}{0.6947} & \textcolor{red}{0.4861} & 0.5029\\ \hline
 \textsf{tv-sports} 	& 0.5156 & 0.6746 & 0.6170 & \textcolor{ForestGreen}{0.7172} & \textcolor{red}{0.5020} & 0.5368\\ \hline
 \textsf{tv-talk} 		& 0.5692 & 0.7142 & 0.6393 & \textcolor{ForestGreen}{0.7364} & \textcolor{red}{0.5299} & 0.5250\\ \hline
 \end{tabular}
\end{table*}

\begin{table*}[h]
\centering
\caption{\label{table:distanceMetrics}
Estimated distances using distribution-based metrics for the proposed algorithm in comparison with the state-of-the-art algorithms using \emph{Scale 1} saliency maps computed using 3DFFT algorithm \cite{long2015saliency}. Note that the highest value in each distance metric is labeled in \textcolor{ForestGreen}{green} and the lowest value in \textcolor{red}{red}. Also, the distance values for the proposed algorithm is shown in \textbf{bold}}
\label{my-label}
\begin{tabular}{|c|l|l|l|l|l|l|l|l|l|l|l|l|}
\hline
\multicolumn{1}{|l|}{} & \multicolumn{4}{c|}{CRCNS}& \multicolumn{4}{c|}{DIEM}& \multicolumn{4}{c|}{AVD}
 \\ \hline
Algorithms             & \multicolumn{1}{c|}{JS}& \multicolumn{1}{c|}{JD}  & \multicolumn{1}{c|}{HI}& \multicolumn{1}{c|}{L2}& \multicolumn{1}{c|}{JS}& \multicolumn{1}{c|}{JD}& \multicolumn{1}{c|}{HI}& \multicolumn{1}{c|}{L2}& \multicolumn{1}{c|}{JS}& \multicolumn{1}{c|}{JD}& \multicolumn{1}{c|}{HI}& \multicolumn{1}{c|}{L2}
\\ \hline
 \hline
\emph{TU} \cite{alshawi15temporal} & 0.34 & 0.68 & 0.64 & 0.25 & 0.16 & 0.32 & 0.30 & 0.12 & 0.33 & 0.66 & 0.60 & 0.24
\\ \hline
\emph{SU} \cite{alshawi15spatial} & 0.14 & 0.29 & 0.32 & 0.12 & 0.05 & 0.09 & 0.13 & \textcolor{ForestGreen}{0.04} & 0.09 & 0.18 & 0.24 & \textcolor{ForestGreen}{0.08}
\\ \hline
\emph{SU+TU} & 0.14 & 0.27 & 0.32 & 0.12 & 0.05 & 0.11 & 0.14 & 0.05 & 0.10 & 0.20 & 0.28 & 0.10
\\ \hline
\emph{STU} & \textcolor{ForestGreen}{\textbf{0.08}} & \textcolor{ForestGreen}{\textbf{0.15}} & \textcolor{ForestGreen}{\textbf{0.24}} & \textcolor{ForestGreen}{\textbf{0.08}} & \textcolor{ForestGreen}{\textbf{0.03}} & \textcolor{ForestGreen}{\textbf{0.05}} & \textcolor{ForestGreen}{\textbf{0.10}} & \textcolor{ForestGreen}{\textbf{0.04}} & \textcolor{ForestGreen}{\textbf{0.06}} & \textcolor{ForestGreen}{\textbf{0.12}} & \textcolor{ForestGreen}{\textbf{0.21}} & \textcolor{ForestGreen}{\textbf{0.08}} 
\\ \hline
\emph{EU} \cite{Fang2014video} & \textcolor{red}{0.51} & \textcolor{red}{1.02} & \textcolor{red}{0.82} & \textcolor{red}{0.46} & \textcolor{red}{0.23} & \textcolor{red}{0.45} & \textcolor{red}{0.36} & \textcolor{red}{0.19} & \textcolor{red}{0.45} & \textcolor{red}{0.91} & \textcolor{red}{0.71} & \textcolor{red}{0.41}
\\ \hline
Baseline & 0.38 & 0.76 & 0.68 & 0.32 & 0.15 & 0.30 & 0.29 & 0.13 & 0.29 & 0.58 & 0.57 & 0.25
\\ \hline
\end{tabular}
\end{table*}

\nocite{Cover91}

\ifCLASSOPTIONcaptionsoff
  \newpage
\fi

\bibliographystyle{IEEEtran}
\bibliography{main}

\end{document}